%% file: main.tex
\documentclass{article}

\usepackage[final]{nips_2018}

\usepackage[utf8]{inputenc} 
\usepackage[T1]{fontenc}    
\usepackage{hyperref}       
\usepackage{url}            
\usepackage{booktabs}       
\usepackage{amsfonts}       
\usepackage{nicefrac}       
\usepackage{microtype}      

\usepackage{algorithm}
\usepackage[noend]{algpseudocode}
\usepackage{caption}
\usepackage{subcaption}
\usepackage{tikz}
\usetikzlibrary{3d,decorations.text,shapes.arrows,positioning,fit,backgrounds}
\usepackage{adjustbox}
\graphicspath{{figures/}}
\PassOptionsToPackage{square}{natbib}
\usepackage{graphicx,wrapfig,lipsum}

\title{Target Driven Visual Navigation with Hybrid Asynchronous Universal Successor Representations}

\author{
  Shamane Siriwardhana\\
  Augmented Human Lab\\
  The University of Auckland\\
  New Zealand \\
  \texttt{shamane@ahlab.org} \\
  \And
  Rivindu Weerasekera\\
  Department of Electrical and Computer Engineering\\
  The University of Auckland\\
  New Zealand \\
  \texttt{rwee015@aucklanduni.ac.nz} \\
  \AND
  Suranga Nanayakkara\\
  Augmented Human Lab\\
  The University of Auckland\\
  New Zealand \\
  \texttt{s.nanayakkara@auckland.ac.nz} \\
}

\begin{document}

\maketitle

\begin{abstract}
Being able to navigate to a target with minimal supervision and prior knowledge is critical to creating human-like assistive agents. Prior work on map-based and map-less approaches have limited generalizability.
In this paper, we present a novel approach, Hybrid Asynchronous Universal Successor Representations (HAUSR), which overcomes the problem of generalizability to new goals by adapting recent work on Universal Successor Representations with Asynchronous Actor-Critic Agents. We 
show that the agent was able to successfully reach novel goals and we were able to quickly fine-tune the network for adapting to new scenes. This opens up novel application scenarios where intelligent agents could learn from and adapt to a wide range of environments with minimal human input.
\end{abstract}

\section{Introduction}

Visual navigation is a core problem in the fields of robotics and machine vision.  
Previous research had used map-based, map-building or map-less approaches~\citep{bonin2008visual,oriolo1995line,borenstein1991vector}.
The first two approaches had been favoured in the past, however, 
they depended on an accurate mapping of the environment and a careful human-guided training phase that limited its generalizability~\citep{filliat2003map}. 

With recent advances in Deep Reinforcement Learning~\citep{mnih2015human, silver2016mastering,silver2017mastering}, map-less navigation~\citep{zhu2017target,zhang2017deep} has seen major advancements.
Reinfocement Learning (RL) systems are composed of an agent that learns by trial \& error. The agent can learn from what worked best for each situation in the past and apply that knowledge when presented with another similar situation. Ideally, the agent transfers its learning among navigation for various destinations or goals within an environment. However, classical RL algorithms struggle to generalize to changing tasks or goals because rewards and value functions, are generally defined in terms of just one goal or target \citep{sutton2011reinforcement}. For example, the value function for a robot navigating to the microwave will be quite different from the one used to navigate to the sofa. 
Previous successful attempts at target-driven navigation using RL have a marked drop in performance when adapting to new goals even though they use models which encourage generalizability~\citep{zhu2017target,zhang2017deep}.

To overcome challenges in generalizability and performance, in this paper, we present a new approach using Hybrid Asynchronous Universal Successor Representations (HAUSR). This is based on the concept of Universal Successor Representations (USR) which are able to learn representations of the environment dynamics that are transferable between different goals. 

We contribute with showing how HAUSR can be adapted to the problem of target driven visual navigation in a complex photo-realistic environmend inside AI2THOR~\citep{kolve2017ai2}. With evaluation of HAUSR in AI2THOR, we demonstrate that an agent was able to successfully reach unseen goals.


\section{Related Work}
 
\subsection{General Value Functions}
One question central to RL is how to learn a feature representation that is scalable and re-uses learned information between tasks.
Designing a value function that is capable of adapting to different tasks, would greatly help generalizability. This line of research builds on a concept called General Value Functions (GVF)~\citep{sutton2011horde}. 
GVFs generalize the concept of the value function to capture not only the goal-oriented semantics of a task but also attempt to capture a more general form of knowledge about the world. 
An extension to this idea, called Universal Value Function Approximators (UVFAs), was introduced by~\cite{schaul2015universal}. The main idea of UVFAs is to represent a large set of optimal value functions by a single, unified function approximator that generalises over both states and goals. Although theoretically sound, learning a good UVFA is a challenging task~\citep{ma2018universal}. 

\subsection{Successor Representations} 
Successor Representation (SR)~\citep{dayan1993improving} emerged from the field of cognitive science and is modeled on how the brain is able to create a reusable predictive map. 
SR was combined with Deep Learning to create Deep Successor Reinforcement Learning (DSR) by ~\cite{kulkarni2016deep} which decoupled reward and environment dynamics. Based on Deep Q-Network (DQN) fundamentals, they were able to learn task-specific features that were able to quickly adapt to distal changes in the reward by fine-tuning only the reward prediction feature vector. 
Transfer in RL was also evaluated on multiple similar tasks by ~\cite{barreto2017successor} who introduced Successor Features (SF). They adapted SR to the continuous domain and were able to show how classic policy improvement can be extended to multiple policies.
Extending ~\cite{barreto2017successor} to Deep Learning, ~\cite{barreto2018transfer} showed how these models could be trained in a stable way.  
For the problem of visual navigation, a SR-based DRL architecture similar to~\cite{kulkarni2016deep} was used by~\cite{zhang2017deep}. Unlike our approach, they showcase 
their solution in simple maze-like environments using DQN as the baseline method, while we use actor-critic methods in a photorealistic simulation environment. 
DQN-based techniques frequently suffer from stability issues when applied to complex problems like large-scale navigation \citep{barreto2018transfer}. 

Universal Successor Representations (USR)~\citep{ma2018universal} are a recent extension to SR. Unlike previous methods which were based on DQN, USR learns a policy directly by modeling it with actor-critic methods. Similar to SR, USR modifies the policy with successor features. DQN-based approaches learn an optimal action-value function indirectly. USR attempts to obtain a General Value Function which can be directly used to obtain an optimal policy. It can be seen as a combination of the SF and UVFA methods as discussed earlier. Unlike methods based on DQN, USR is able to directly optimize an agent to learn multiple tasks simultaneously. However, USR has not been tested on high-dimensional complex problems. This paper shows how USR could be adapted to the problem of target driven visual navigation in a photorealistic environment. 

\subsection{Target-Driven Navigation}
There were several attempts at solving the visual navigation with DRL.
The most relevant to our approach used a deep siamese actor critic agent that shares parameters across multiple goals~\citep{zhu2017target}. They used the asynchronous advantage actor critic (A3C)~\citep{mnih2016asynchronous} algorithm and were able to learn multiple goals simultaneously by feeding in both state and goal features as inputs to train across multiple goals simultaneously.  Although their approach intuitively resembles General Value Function approximation, their results show a marked decrease in performance when navigating to targets which the agent has not seen during training. Unlike our approach, they did not use any form of successor representations for generalizations in their architecture.   
Many RL algorithms are trained in simulated environments that are usually simplified lower dimensional versions of reality~\citep{beattie2016deepmind}. As a result, the agent does not use rich visual information to make decisions. However with the development of photo-realistic simulated environments like AI2THOR \citep{kolve2017ai2}, recent work have been able to use rich visual information and successfully transfer knowledge from a simulated environment to the real world \citep{zhu2017target}.

\section{Background}

\subsection{Reinforcement Learning}

We formalize the goal-directed navigation task as a Markov Decision Process (MDP). The transition probability $p(s_{t+1}|s,a)$ defines the probability of reaching the next state $s_{t+1}$ when action $a_t \in \mathcal{A}$ is taken in state $s_t \in \mathcal{S}$. For any goal $g \in \mathcal{G}$ (very often $\mathcal{G} \subseteq \mathcal{S}$), we define a pseudo-reward function $r_g(s_t, a_t, s_{t+1}) \in \mathbb{R}$ and a pseudo-discount function $\gamma_g(s) \in [0, 1]$ (for terminal state, $\gamma_g =0$). For any policy 
$\pi(a_t|s_t)$
, the General Value Function \citep{sutton2011horde} can be defined as:

\begin{equation}
    \label{eq:Value}
    V_{g}^{\pi}(s) = \mathbb{E}^\pi \left [ 
    \sum_{t=0}^{\infty}
    r_g(s_t,a_t,s_{t+1})\left . \prod_{k=0}^{t}\gamma_g (s_k)  \right | s_0=s  \right ]
\end{equation}

We assume for any goal $g$, there exists an optimal value function $V_{g}^{*}(s) = V_{g}^{\pi^*_g}(s)$ evaluated according to a goal oriented optimal policy $\pi^{*}_{g}$. The aim of policy learning is to find the optimal policy $\pi^*$ that maximizes the future discounted rewards starting from $s_0$ and following $\pi^*$. 

To generalize over the goal space $\mathcal{G}$, the agent needs to learn multiple optimal policy and optimal value functions for navigating to each goal. Each goal is considered a new task for the agent and it should be able to quickly adapt to these new tasks and learn $V_{g}^{\pi^*_g}(s)$ and $\pi^{*}_{g}$.

\subsection{Asynchronous Advantage Actor Critic (A3C)}

Asynchronous advantage actor-critic (A3C) is an on-policy policy search method introduced by~\cite{mnih2016asynchronous} based on actor-critic methods~\citep{witten1977adaptive,barto1983neuronlike}. A3C is a fast, robust, scalable and has achieved state of the art results in the video game domain~\citep{mnih2016asynchronous}.

A3C maintains estimates of both a policy $\pi(a_t|s_t)$ and the value
function $V(s_t)$. The agent uses the value function (the critic) to update it's policy (the actor) by training multiple threads in parallel and asynchronously updating a shared set of model parameters. It has been shown that the parallel threads are able to stabilize each other. In practice we can learn both policy and value functions in a single neural network with multiple heads.

An update is performed by calculating 
$\nabla_{\theta_\pi} \log \pi(a_t|s_t) A(s_t,a_t)$
 where $A$ is the advantage term. The advantage term can be thought of as quantifying how much better an action turned out to be than expected. For a state, the advantage can be estimated by the difference between the discounted rewards of an episode and the value of that state. The full update can therefore be written as:
\begin{equation}
    \nabla_{\theta_\pi} \log \pi(a_t|s_t;\theta_\pi) (R_t-V(s_t;\theta_V))+\beta\nabla_{\theta_\pi} H(\pi(s_t;\theta_\pi))
\end{equation}
$H$ is an entropy terms which encourages exploration of the agent controlled by hyperparameter $\beta$.

\input{figure2o.tex}

\subsection{Universal Successor Representations (USR)}

A key idea in applying SR to deep architectures is being able to decouple and approximate the reward function $r_g(s_t,a_t,s_{t+1})$ as a linear combination of learned state features $\phi(s_t,a_t,s_{t+1})$ and a reward weight vector $\omega(g)$~\citep{kulkarni2016deep,barreto2017successor}.
\begin{equation}
    r_g \approx \phi(s_t,a_t,s_{t+1};\theta_\phi)^\top \omega(g_t;\theta_\omega)\\
    ~\approx \phi(s_{t+1};\theta_\phi)^\top \omega(g_t;\theta_\omega)
\label{eq:rg}
\end{equation}
where $\theta_\phi$ and $\theta_\omega$ are learnt parameters of a function approximator such as a neural network. In this case, equation \ref{eq:Value} can be rewritten as

\begin{equation}
    \label{eq:ValueUSR}
    \resizebox{\linewidth}{!}{$
        V_{g}^{\pi}(s)  = \mathbb{E}^\pi \left [ 
        \sum_{t=0}^{\infty}
        \phi(s_t,a_t,s_{t+1};\theta_\phi)\left . \prod_{k=0}^{t}\gamma_g (s_k)  \right | s_0=s  \right ]^\top \omega(g_t;\theta_\omega) 
      = \psi^\pi(s_t,g_t;\theta_\psi)^\top \omega(g_t;\theta_\omega)
    $}
\end{equation}

where $\psi^\pi(s_t,g_t)$ is defined as the USR of state $s_t$~\citep{ma2018universal}. Intuitively,  $\psi^\pi(s_t,g_t)$ can be thought of as the expected future state occupancy.

USR makes it easier for us to transfer knowledge between goals. 
If $\omega(g_t;\theta_\omega)$ can be effectively computed for any $g_t$, we can compute a value function and then an optimal policy for any goal. Learning the USR can then be done in the same way as the value function update using the following Bellman update:
\begin{equation}
    \psi^\pi(s_t,g_t;\theta_\psi) = \mathbb{E}^\pi[\phi(s_t,a_t,s_{t+1})+\gamma_g(s)\psi^\pi(s_{t+1},g_t;\theta_\psi)]
\label{eq:psi}
\end{equation}

\begin{wrapfigure}{r}{5.5cm}
\centering
\begin{adjustbox}{width=0.3\textwidth}
\begin{tikzpicture}
    \foreach \X [count=\i] in {66,28}
    {
        \node [inner sep=0pt] at ({-(\i*4)-1},0) {\includegraphics[width=0.28\textwidth]{states/color/out_\X}}; 
    }
    \foreach \X [count=\i] in {71,57}
    {
        \node [inner sep=0pt] at ({-(\i*4)-1},3) {\includegraphics[width=0.28\textwidth]{states/color/out_\X}}; 
    }
\end{tikzpicture}
\end{adjustbox}
\caption{States coming from the AI2THOR environment are $\textbf{300}\times\textbf{400}$ RGB images}
\label{fig:colorStates}
\end{wrapfigure}

\section{Problem Formulation}

The objective of our target driven navigation agent is to learn a stochastic policy function $\pi(s_t,g_t)$ where $s_t$ is the representation of the current state and $g_t$ is a representation of the target state. The output of policy $\pi$ would be a probability distribution over actions $\mathcal{A}$. In our formulation, the agent has four discrete actions: 1) to move forward 0.5m, 2) move backward 0.5m, 3) turn left or 4) turn right. 
We use the ``bathroom 02'' which contains $180$ unique states in the photo realistic 3D AI2THOR~\citep{kolve2017ai2} environment to train our policy and test the transferability of the learned policy to other novel goals.

Each scene is divided into $0.5m \times 0.5m$ grids similar to a grid world environment. The state representation coming from the AI2THOR environment is a $300\times400$ raw colour pixel image~\ref{fig:colorStates}. We scale this down to a $110\times110$ gray-scale image and concatenate four consecutive frames to create our state vector $s_t$. 

The agent receives a reward of $+1$ for reaching the goal state and $-0.01$ for each time-step. The objective of our agent is to navigate to a goal location in a minimal number of steps.  After training, the agent should be able to generalize to novel goal locations which were not used to train our model.

\subsection{Challenges in merging  USR with A3C}

USR and actor-critics methods have previously only been applied to simple maze-like environments. We found applying USR under the baseline method of A3C directly to the complex problem of visual navigation to be unstable due to a few reasons. 

First, learning a good state embedding $\phi$ (Equation \ref{eq:psi}) using just an auto-encoder as proposed by~\cite{ma2018universal} in a complex environment (like AI2THOR) was a challenge. The state representation is a crucial part required for the agent to gain an understanding of it's environment. Mistakes in the state representation caused degraded performance of the agent.

Second, we found that it was not easy to accurately predict the reward prediction vector ($\omega(g_t)$ in Equation \ref{eq:rg}). This vector and state representations are in turn are used in Equation~\ref{eq:ValueUSR} and help us approximate the advantage function and scalar reward. Unlike ad-hoc rewards, the predicted scalar rewards using Equation \ref{eq:psi} have wide variations. This is especially the case when training for multiple goals simultaneously. In the next section we describe how we address these challenges.  

In the next section we will go over our Hybrid Asynchronous Universal Successor Representations (HAUSR) architecture. To our knowledge, this is the first time USR has been combined with A3C for target driven visual navigation.

\section{Hybrid Asynchronous Universal Successor Representations (HAUSR)}

\subsection{Network Architecture}
Different from previous methods which used DQN as the baseline method with SR to improve the generalizability and transfer learning capabilities, we used a deep actor critic method and extend USR to work with A3C. This has the advantage of opening up our algorithm to be used in real world sceneraios due to A3C's scalability~\citep{mnih2016asynchronous}.
Our architecture is also more stable and scalable than the vanilla USR network as proposed by~\cite{ma2018universal}.
The three main networks of our architecture are shown in Figure \ref{fig:network}. 
The state representation and reward prediction networks (Figure \ref{fig:network}a) were trained before the reinforcement learning network (Figure \ref{fig:network}b) was trained.

\begin{wrapfigure}{r}{6cm}
\centering
\begin{adjustbox}{width=0.4\textwidth}
\begin{tikzpicture}
    \foreach \X [count=\i] in {1,2,3,4}
    {
        \begin{scope}
            \clip [rounded corners=.5cm] ({-(\i*2)+1},1) rectangle coordinate (real\i) ({-(\i*2)-1},3);
            \node [inner sep=0pt] at (real\i) {\includegraphics[width=0.13\textwidth]{states/recons/out_real3_\X}}; 
        \end{scope}
        \begin{scope}
            \clip [rounded corners=.5cm] ({-(\i*2)+1},0.5) rectangle coordinate (gen\i) ({-(\i*2)-1},-1.5);
            \node [inner sep=0pt] at (gen\i) {\includegraphics[width=0.13\textwidth]{states/recons/out_gen3_\X}}; 
        \end{scope}
    }
\end{tikzpicture}
\end{adjustbox}
\caption{The top row shows some of the states in the simulator. The bottom row shows the reconstruction of those states generated by feeding the learnt state representation $\phi$ through the auto-encoder decoder}
\label{fig:reconstruction}
\end{wrapfigure}

\subsubsection{State Representation Network}

The goal of the state representation network was to generate $\phi$ which encodes useful dynamics of the state that would later be used to approximate the successor feature representations. 
Instead of training $\phi$ with just an auto-encoder loss, 
we used a new architecture with auxiliary losses to make a more robust state representation. 
We added auxiliary branches in the form of a foreword dynamics loss and an inverse dynamics loss, in addition to the autoencoder reconstruction loss (as shown in Figure \ref{fig:network}a). Forward and inverse dynamics have been widely discussed in literature and have shown to improve state representations for DRL tasks~\citep{christiano2016transfer,pathak2018zero}. Both autoencoder reconstruction and forward dynamics losses use a mean squared error loss function, while the inverse dynamics loss uses a cross-entropy loss. 

In the early stages of training, states were sampled from exploration of the agent with a randomly initialized policy. In each interaction with the environment we collected roll-outs consisting of $s_t,a_t,s_{t+1}...$.
Using these rollouts, we were able to train an informative bottleneck layer using Algorithm~\ref{alg:1} and generate $\phi$.
Figure \ref{fig:reconstruction} shows some of the reconstructed images of different states using $\phi$.

\subsubsection{Reward Prediction Network}

The goal of the reward prediction network was to train the reward prediction vector $\omega_g$. This vector should be able to transform the state representation $\phi$ into a scalar reward as
\begin{equation}
    r_g \approx \omega \cdot \phi^\top
\end{equation}

We used a three layer Convolutional Neural Network (CNN) which took in the target image as input as generated $\omega$ as shown in Figure \ref{fig:network}a. The predicted rewards should be either $-0.01$ if the agent had not reached the goal or $+1$ if the agent had reached the goal. Generating $\omega$ with this simple network gave us the ability to quickly obtain the reward prediction vector for different goals. First, we pre-trained this network with the collected roll-outs and then trained simultaneously with the A3C agent.

\subsubsection{Reinforcement Learning Network}

The final network was our Actor-Critic agent network (Figure \ref{fig:network}b) which implements our A3C agent with successor features.  

We created a new advantage function in order to achieve stability during training.
The advantage function is normally the driving force for policy improvement and is based on ad-hoc rewards (like a small penalty for every step and +1 for reaching the goal). Advantage is calculated as the difference between the discounted reward and the value of that state.
An ad-hoc scaler rewards are very useful for optimizing agents in complex environments, especially with parallel architectures like A3C. Unlike DQN, A3C is stable out of the box and generally require less hyperparameter tuning.

In our work, we combined the normal A3C architecture with USR by calculating two advantage functions. One advantage function was calculated with ad-hoc scaler rewards (like in traditional A3C), while the other was calculated with the output from the reward prediction network (which took the successor features into account).
We modified the conventional advantage and return functions in the A3C agent with USR as follows:

\begin{equation}
    \hat{A}_t^{\psi} = \left[\phi(s_t)+\gamma\psi(s_{t+1},g_t) - \psi(s_t,g_t)\right]^\top \omega_g
\end{equation}
\begin{equation}
    \hat{A}_t^{V} = \sum_{k=0}^{\infty} \gamma^k r_{t+k} - V_t 
\end{equation}

We took a weighted sum of the two advantages with the hyperparameter $\lambda$, determining the contribution from successor feature rewards. As a result, the agent was able to update its policy parameters not only with respect to ad-hoc rewards but also took into account the state and environment dynamics. We believe this is key to learning better feature representations that are able to generalize across goals. 

In the RL network, we sent both the goal image and current state through a shared three layer siamese CNN. We then concatanated the two output vectors from the CNN and passed it through fully connected layers. The output of the network was a value function~$V$, policy $\pi$ and a Universal Successor Representation Approximator~$\psi$. It was trained as described in Algorithm~\ref{alg:2}. 
The model can be easily extended to multiple environments by creating branches after the convolutional layers of the network. 

\subsection{Training Protocol}

Constructing a stable training architecture was very important to applying Deep Reinforcement Learning to high dimensional problems, such as visual navigation. 

Both the return and the advantage functions of our network depend on the reward prediction vector ($\omega_g$) and the state representation vector ($\phi$). Conventional A3C agents rely on a stable reward generating mechanism. Obtaining perfect state representation and reward prediction vectors was highly unlikely due the noise and complexities in the real world. 
During training, we initially set $\lambda$ to a very low value to prioritize learning of the more stable value function. After $5k$ iterations with 32 agents, we increased $\lambda$ to $0.001$. After we increased $\lambda$, the value loss also increased, providing evidence that the SR loss had an effect on training. 
The combined loss function consisted of the policy loss, value function loss, successor feature approximation loss and an entropy loss~\citep{williams1991function}.

\begin{equation}
    \mathcal{L}_{total} = \lambda \mathcal{L}_{\psi}  + \mathcal{L}_{V} + \mathcal{L}_\pi + \mathcal{L}_{H}
\end{equation} 

It is our understanding that this is the first time that the conventional value loss was used in conjunction with successor features. The addition of auxiliary losses to help training has been done before~\citep{jaderberg2016reinforcement} and we believe this concept will be key to creating successful real world RL agents in the future.

\section{Evaluation}

\begin{wrapfigure}{r}{6cm}
    \centering
    \includegraphics[width=0.45\textwidth]{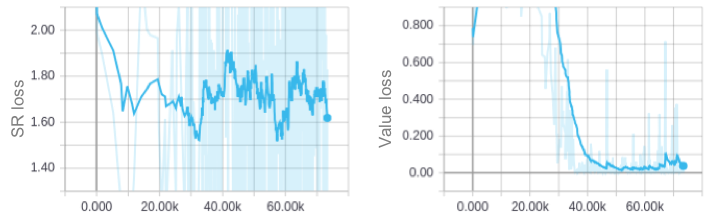}
    \caption{Shows how our SR loss and Value Loss stabilize over time}
    \label{fig:training}
\end{wrapfigure}

Figure~\ref{fig:training} shows the n-step temporal difference error with respect to the conventional Value function as well as the Successor Feature approximator. \input{figure1.tex}
We can see how these stabilize over time. As evident in the graph, the SR loss does not reach zero but this is sufficient to give generalization abilities to our agent. It also provides evidence for why we train in a hybrid manner and still use the traditional Value loss in our training. We didn't compare our model with a baseline model of A3C since we were more concerned about creating a stable training architecture for combining A3C with USR in a high dimensional complex problem. 

We tested the capability of our model for transfer learning and generalization by conducting two experiments. One to test the generalizability of the agent without retraining and the other to test how quickly we can fine tune the agent for novel goals. For both experiments, we trained our model until convergence in the bathroom environment of AI2THOR with five goal locations.

\subsection{Generalization to novel goals}

\begin{figure}
    \centering
    \includegraphics[width=0.95\textwidth]{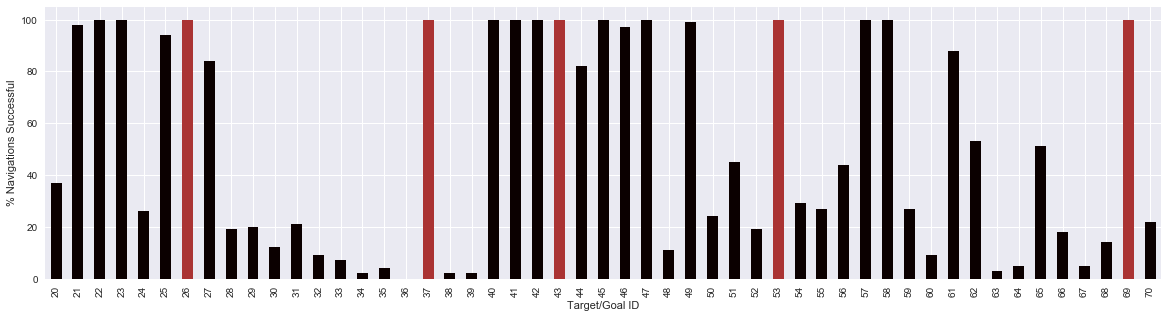}
    \caption{Success rate of agent navigating to goal for all tested states. The red highlighted states are the targets obtained from the AI2THOR simulator and used for training. The black states were not used for training. The agent was able to successfully generalize it's learnings and navigate to a large number of new goals. Note: the states are not necessarily in order of closeness}
    \label{fig:results1}
\end{figure}

We checked the generalizability of our agent to new tasks \textit{without} fine-tuning. 
When generalizing to new goals, the most important factor to consider was if the agent was successful in reaching the new goal. Therefore, following \citep{zhu2017target}, we used the success rate of reaching the goal as a metric. For each novel goal, we ran $100$ episodes of the agent with each starting from a random initial state. If the agent reached the goal within $500$ steps, we considered it to be a successful episode. 
We tested our agent's ability to navigate to $50$ goal states in the same environment. Figure \ref{fig:results1} shows the success rate of our agent in reaching these goals. 
Note that we do not fine tune our network for these new goals.  
\subsection{Transfer Learning Ability}
\begin{wrapfigure}{r}{5cm}
    \centering
    \includegraphics[width=0.3\textwidth]{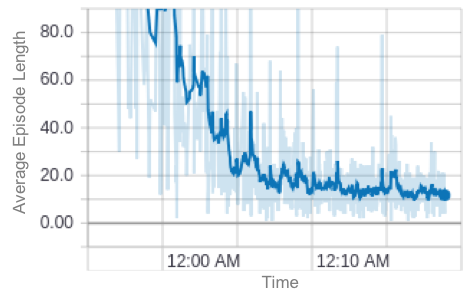}
    \caption{Transfer Learning: the agent is able to learn to navigate to novel goals it was previously unable to reach in under $15$ minutes of fine-tuning}
    \label{fig:transfer}
\end{wrapfigure}

Transfer learning has been extensively studied in the supervised learning domain and leads to fast training on new tasks~\citep{sharif2014cnn}. If a model is able learn a rich representation of a problem, that knowledge should be useful when learning new tasks. This enables us to have an agent that can quickly adapt to new tasks. 
We tested our network's transfer learning ability by fine-tuning the network for new goals. 
Figure~\ref{fig:transfer} shows that our agent is able to very quickly learn to navigate to goals that it previously not able to reach. The number of steps taken to reach a new goal dramatically decreases and the agent is able learn in under $15 minutes$.

\subsection{Future Work}

In the recent past, DRL has been very effective at solving complex tasks \citep{silver2016mastering,mnih2015human}. Real world navigation is still an open problem in DRL. New breakthrough algorithms are frequently only tested in simple toy environments. Scalability is another factor that needs to be considered when applying DRL to complex domains. We identified that merging new concepts into complex domains is challenging and a key research question is how these ideas can be implemented with stable architectures. We hope that our initial attempt at merging two novel concepts leads to more research in creating fine-tuned architectures that are able to perform reliably in real-world applications. 

\section{Conclusion}

We show the applicability of Universal Sucessor Representations to the complex domain of target driven visual navigation in a photorealistic environment. We present a new architecture (HAUSR) that is able to create rich successor representations and train asynchronously with A3C. Despite encouraging results, there are several opportunities to exploit hybrid Successor Representation based approaches to achieve higher generalizability and transferrability of RL agents. 


\bibliographystyle{plainnat}
\bibliography{library}

\begin{center}
\textbf{\large Appendix}
\end{center}
\input{algorithms.tex}

\end{document}

%% file: figure2o.tex
\tikzset{pics/box3d/.style args={color #1 thick #2 tall #3 deep #4 text #5}{
        code={
            \draw[gray,ultra thin,fill=#1]  (0,0,0) 
                coordinate(-front-bottom-left) to ++ (0,#3,0) 
                coordinate(-front-top-right) --++ (#2,0,0) 
                coordinate(-front-top-right) --++ (0,-#3,0) 
                coordinate(-front-bottom-right) -- cycle;
            \draw[gray,ultra thin,fill=#1] (0,#3,0)  --++ (0,0,#4) 
                coordinate(-back-top-left) --++ (#2,0,0) 
                coordinate(-back-top-right) --++ (0,0,-#4)  -- cycle;
            \draw[gray,ultra thin,fill=#1!80!black] (#2,0,0) --++ (0,0,#4) 
                coordinate(-back-bottom-right) --++ (0,#3,0) --++ (0,0,-#4) -- cycle;
            \path[gray,decorate,decoration={text effects along path,text={#5}}] (#2/2,{2+(#3-2)/2},0) -- (#2/2,0,0);
        }
    }
}

\begin{figure*}[ht]
\centering
\begin{subfigure}[b]{0.45\textwidth}
    \centering 
    \begin{adjustbox}{width=\textwidth}

\begin{tikzpicture}[x={(1,0)},y={(0,1)},z={({cos(60)},{sin(60)})},font=\sffamily\small,scale=1]

    
    \draw (0,1.75,0) node[minimum height=2cm,minimum width=2cm,draw,ultra thick,rounded corners,fill=white!90!blue,scale=2,opacity=0.7] (st1) {$s_{t+1}$};
    \draw (0,10.75,0) node[minimum height=2cm,minimum width=2cm,draw,ultra thick,rounded corners,fill=white!70!blue,scale=2,opacity=1] (st) {$s_{t}$};
    
    \draw[-latex,ultra thick] (st) to [bend right=20] (st1) node[fill=red!20,circle,draw,ultra thick,scale=2] at (0,6,0) {$a_t$};
    
    
    \foreach \X in {0.85,0.9,0.95,1.0}
    {
        \draw[canvas is yz plane at x = \X, transform shape, draw = gray, fill =gray, opacity = 0.5] (0,0) rectangle (2.5,1);
        \draw[canvas is yz plane at x = \X, transform shape, draw = gray, fill =gray, opacity = 0.5] (9,0) rectangle (11.5,1);
    }
    \coordinate (c1a) at ({2.4},{9},{0});
    \draw pic (c1-1) at (c1a) {box3d= color {white!70!blue} thick {0.1} tall {2.5} deep {1} text { }};
    \coordinate (c2) at ({2.4},{0},{0});
    \draw pic (c2-1) at (c2) {box3d= color {white!90!blue} thick {0.1} tall {2.5} deep {1} text { }};
    \coordinate (c1b) at ({2.6},{9.1},{0});
    \draw pic (c1-2) at (c1b) {box3d= color {white!70!blue} thick {0.1} tall {2.3} deep {0.8} text { }};
    \coordinate (c2b) at ({2.6},{0.1},{0});
    \draw pic (c2-1) at (c2b) {box3d= color {white!90!blue} thick {0.1} tall {2.3} deep {0.8} text { }};
    \coordinate (c1c) at ({2.8},{9.2},{0});
    \draw pic (c1-2) at (c1c) {box3d= color {white!70!blue} thick {0.1} tall {2.1} deep {0.6} text { }};
    \coordinate (c2c) at ({2.8},{0.2},{0});
    \draw pic (c2-1) at (c2c) {box3d= color {white!90!blue} thick {0.1} tall {2.1} deep {0.6} text { }};
    \node at (3,9,0) (lt2) {$\theta_\phi$};
    
    \draw[black,ultra thick,-latex] (3,10.5,0) -- (4,10.5,0);
    \draw[black,ultra thick,-latex,opacity=0.5] (3,1.5,0) -- (4,1.5,0);
    
    \filldraw[black!90,ultra thick,rounded corners,fill=blue!30] (4,9,0) rectangle (4.9,12,0) node[black!100,pos=0.5,rotate=90,scale=1.2] {$\phi(s_{t})$}; 
    \filldraw[black!90,ultra thick,rounded corners,fill=blue!30,opacity=0.5] (4,0,0) rectangle (4.9,3,0) node[black!100,pos=0.5,rotate=90,scale=1.2] {$\phi(s_{t+1})$}; 
    
    \draw[black,ultra thick,-latex] (4.9,10.5,0) -- (6,10.5,0) -- (6,12.5,0) -- (8.5,12.5,0);
    \draw[black,ultra thick,-latex] (4.9,10.5,0) -- (6,10.5,0) -- (6,8,0) -- (7.3,8,0);
    \draw[black,ultra thick,-latex] (4.9,10.5,0) -- (6,10.5,0) -- (6,3,0) -- (7,3,0);
    
    \filldraw[black!90,ultra thick,rounded corners,fill=green!30] (7,1,0) rectangle (7.7,4,0) node[black!50,pos=0.5,rotate=90] {Concat-1024};
    \draw[black,ultra thick,-latex,dashed] (4.9,1.5,0) -- (6,1.5,0) -- (6,2,0) -- (7,2,0);
    \filldraw[black!90,ultra thick,rounded corners,fill=green!30] (8.5,1.25,0) rectangle (9,3.75,0) node[black!50,pos=0.5,rotate=90] {FC-512};
    \draw[black,ultra thick,-latex] (7.7,2.5,0) -- (8.5,2.5,0) (9,2.5,0) -- (10,2.5,0);
    \node[fill=red!20,circle,draw,ultra thick,scale=2,opacity=0.7] at (10.7,2.5,0) {$a_t$};
    
    \filldraw[black!90,ultra thick,rounded corners,fill=purple!30] (7.3,6,0) rectangle (8,9,0) node[black!50,pos=0.5,rotate=90] {Concat-1024};
    \filldraw[black!90,ultra thick,rounded corners,fill=purple!30,opacity=0.7] (6.3,6.25,0) rectangle (6.8,7.75,0) node[black!50,pos=0.5,rotate=90] {FC-512};
    \draw[black!50!red,ultra thick,-latex] (0.7,6,0) -- (3,6,0)--(3,7,0)--(6.3,7,0) (6.8,7,0) -- (7.3,7,0);
    
    \coordinate (c3) at ({8.5},{6},{0.4});
    \draw pic (f1-3) at (c3) {box3d= color {white!70!purple} thick {0.1} tall {1.9} deep {0.4} text { }};
    \coordinate (c2) at ({8.8},{6},{0.2});
    \draw pic (f1-2) at (c2) {box3d= color {white!70!purple} thick {0.1} tall {2.1} deep {0.6} text { }};
    \coordinate (c1) at ({9.1},{6},{0});
    \draw pic (f1-1) at (c1) {box3d= color {white!70!purple} thick {0.1} tall {2.3} deep {0.8} text { }};
    \draw[black,ultra thick,-latex] (8,7.5,0) --(8.5,7.5,0);
    \filldraw[black!90,ultra thick,rounded corners,fill=blue!30,opacity=0.4] (10,6.5,0) rectangle (12,8.5,0) node[black!100,pos=0.5,scale=1.2] {$s_{t+1}$}; 
    \foreach \X in {10.15,10.2,10.25,10.3}
    {
        \draw[canvas is yz plane at x = \X, transform shape, draw = gray, fill =gray, opacity = 0.5] (6.75,0) rectangle (7.75,0.5);
    }
    
    \coordinate (ae3) at ({8.5},{11},{0.2});
    \draw pic (f1-3) at (ae3) {box3d= color {white!70!orange} thick {0.1} tall {1.9} deep {0.4} text { }};
    \coordinate (ae2) at ({8.8},{10.9},{0.1});
    \draw pic (f1-2) at (ae2) {box3d= color {white!70!orange} thick {0.1} tall {2.1} deep {0.6} text { }};
    \coordinate (ae1) at ({9.1},{10.8},{0});
    \draw pic (f1-1) at (ae1) {box3d= color {white!70!orange} thick {0.1} tall {2.3} deep {0.8} text { }};
    \filldraw[black!90,ultra thick,rounded corners,fill=blue!30,opacity=0.6] (10,11.5,0) rectangle (12,13.5,0) node[black!100,pos=0.5,scale=1.2] {$s_{t}$}; 
    \foreach \X in {10.15,10.2,10.25,10.3}
    {
        \draw[canvas is yz plane at x = \X, transform shape, draw = gray, fill =gray, opacity = 0.5] (11.75,0) rectangle (12.75,0.5);
    }
    \node[black,ultra thick,anchor=west] at (12,.5,0) (idst) {};
    \node[black,ultra thick,anchor=west] at (12,5.5,0) (fdst) {};
    \node[black,ultra thick,anchor=west] at (12,10.5,0) (aest) {};

    \node[black!40] at (-2,14,0) (lt1) {MDP};
    \node[black!40,anchor=west] at (6,14,0) (ael) {Autoencoder Loss};
    \node[black!40,anchor=west] at (6,9.5,0) (fdl) {Forward Dynamics Loss};
    \node[black!40,anchor=west] at (6,4.5,0) (idl) {Inverse Dynamics Loss};
    \node[black!40,anchor=west] at (0,-0.9,0) (rp) {Reward Prediction Network};
    \node[black,ultra thick,anchor=west] at (9,-5,0) (rpe) {};
    \begin{scope}[on background layer]
        \node[rounded corners,fill=gray!10, fit=(lt1)(st) (st1)]{};
        \node[rounded corners,fill=gray!10, fit=(ael) (aest)]{};
        \node[rounded corners,fill=gray!10, fit=(fdl) (fdst)]{};
        \node[rounded corners,fill=gray!10, fit=(idl) (idst)]{};
        \node[rounded corners,fill=gray!10, fit=(rp) (rpe)]{};
    \end{scope} 
    

    \draw[black,ultra thick,dotted] (-3,-0.5,0)--(13,-0.5,0);

    \node[scale=2] at (1.5,-3,0) {$g$};
    \foreach \X in {1.85,1.9,1.95,2.0}
    {
        \draw[canvas is yz plane at x = \X, transform shape, draw = gray, fill =gray, opacity = 0.5] (-2,0) rectangle (-4.5,1);
    }
    \coordinate (c2) at ({2.2},{-4.5},{0});
    \draw pic (c2-1) at (c2) {box3d= color {white!80!green} thick {0.1} tall {2.5} deep {1} text { }};
    \coordinate (c2b) at ({2.4},{-4.4},{0});
    \draw pic (c2-1) at (c2b) {box3d= color {white!80!green} thick {0.1} tall {2.3} deep {0.8} text { }};
    \coordinate (c2c) at ({2.6},{-4.3},{0});
    \draw pic (c2-1) at (c2c) {box3d= color {white!80!green} thick {0.1} tall {2.1} deep {0.6} text { }};
    \filldraw[black!90,ultra thick,rounded corners,fill=red!30] (4,-4.5,0) rectangle (4.7,-1.5,0) node[black!90,pos=0.5,rotate=90] {$\omega_{g}$}; 
    \filldraw[black!90,ultra thick,rounded corners,fill=yellow!20] (8,-3.5,0) rectangle (9,-2.5,0) node[pos=0.5] {$r_g$};
    \node[black!90,ultra thick,fill=gray!20,circle,draw,scale=2] at (6.5,-3,0) {$\cdot$};
    \draw[black,ultra thick,-latex] (3,-3,0) -- (4,-3,0);
    \draw[black,ultra thick,-latex] (4.7,-3,0)--(6,-3,0);
    \draw[black,ultra thick,-latex] (7,-3,0) --(8,-3,0);
    \draw[black,ultra thick,-latex,dashed] (4.9,1,0)--(5.5,1,0)--(5.5,-1,0)--(6.5,-1,0)--(6.5,-2.5,0);
\end{tikzpicture}
    \end{adjustbox}
    \caption{State Representation and Reward Prediction Networks}
    \label{fig:rps}
\end{subfigure}
\begin{subfigure}[b]{0.53\textwidth}
    \centering
\begin{adjustbox}{width=\textwidth}
    \begin{tikzpicture}[x={(1,0)},y={(0,1)},z={({cos(60)},{sin(60)})},font=\sffamily\small]
    \node at (0,-4,0) {};
    \node[scale=2] at (0.5,1.5,0) {$g$};
    \node[scale=2] at (0.5,6.5,0) {$s_t$};
    \foreach \X in {0.85,0.9,0.95,1.0}
    {
        \draw[canvas is yz plane at x = \X, transform shape, draw = gray, fill =gray, opacity = 0.5] (0,0) rectangle (2.5,1);
        \draw[canvas is yz plane at x = \X, transform shape, draw = gray, fill =gray, opacity = 0.5] (5,0) rectangle (7.5,1);
    }
    \coordinate (c1a) at ({1.2},{5},{0});
    \draw pic (c1-1) at (c1a) {box3d= color {white!70!blue} thick {0.1} tall {2.5} deep {1} text { }};
    \coordinate (c2) at ({1.2},{0},{0});
    \draw pic (c2-1) at (c2) {box3d= color {white!80!green} thick {0.1} tall {2.5} deep {1} text { }};
    \coordinate (c1b) at ({1.4},{5.1},{0});
    \draw pic (c1-2) at (c1b) {box3d= color {white!70!blue} thick {0.1} tall {2.3} deep {0.8} text { }};
    \coordinate (c2b) at ({1.4},{0.1},{0});
    \draw pic (c2-1) at (c2b) {box3d= color {white!80!green} thick {0.1} tall {2.3} deep {0.8} text { }};
    \coordinate (c1c) at ({1.6},{5.2},{0});
    \draw pic (c1-2) at (c1c) {box3d= color {white!70!blue} thick {0.1} tall {2.1} deep {0.6} text { }};
    \coordinate (c2c) at ({1.6},{0.2},{0});
    \draw pic (c2-1) at (c2c) {box3d= color {white!80!green} thick {0.1} tall {2.1} deep {0.6} text { }};
    \node[draw,double arrow, orange,fill=orange!30,rotate=90] at (1.4,3.5,0.6) {Shared};
    
    \draw[black,ultra thick,-latex] (2,6.5,0) -- (3,6.5,0) -- (3,4.5,0) -- (4,4.5,0);
    \draw[black,ultra thick,-latex] (2,1.5,0) -- (3,1.5,0) -- (3,3.5,0) -- (4,3.5,0);
    
    \filldraw[black!90,ultra thick,rounded corners,fill=red!30] (4,2,0) rectangle (4.7,6,0) node[red!50,pos=0.5,rotate=90] {FC-1024}; 
    \draw[black,thick,-latex] (4.7,4,0) -- (5,4,0);
    \filldraw[black!90,ultra thick,rounded corners,fill=red!30] (5,2.5,0) rectangle (5.7,5.5,0) node[red!50,pos=0.5,rotate=90] {FC-512}; 
    
    \draw[black,ultra thick,-latex] (5.7,4,0) -- (7,4,0);
    
    \filldraw[black!90,ultra thick,rounded corners,fill=purple!30] (7,2.5,0) rectangle (7.7,5.5,0) node[purple!50,pos=0.5,rotate=90] (1512){FC-512};
    
    \draw[black,ultra thick,-latex] (7.7,4.0,0) -- (8,4,0) -- (8,7,0) -- (9.4,7,0);
    \draw[black,ultra thick,-latex] (7.7,4,0) -- (8,4,0) -- (8,4,0) -- (11,4,0);
    \draw[black,ultra thick,-latex] (7.7,4,0) -- (8,4,0) -- (8,1,0) -- (9.4,1,0);
   
    \filldraw[black!90,ultra thick,rounded corners,fill=green!30] (9.4,0,0) rectangle (9.9,2,0) node[green!50,pos=0.5,rotate=90] (GN512){FC-512};
    \draw[black,ultra thick,-latex] (9.9,1,0) -- (11,1,0);
    \filldraw[black!90,ultra thick,rounded corners,fill=purple!20] (11,1.5,0) rectangle (12,0.5,0) node[pos=0.5] {$V$};
   
    \node[rectangle,draw,black!80,fill=red!30,scale=0.75,rounded corners] (A1) at (11.5,4,0) {$a_1$};
    \node[rectangle,draw,black!80,fill=red!30,scale=0.75,rounded corners,below=1pt of A1] (A2) {$a_2$};
    \node[rectangle,draw,black!80,fill=red!30,scale=0.75,rounded corners,below=1pt of A2] (A3) {$a_3$};
    \node[rectangle,draw,black!80,fill=red!30,scale=0.75,rounded corners,below=1pt of A3] (A4) {$a_4$};
    \node[black,ultra thick,scale=1.3,above=1pt of A1] (pi) {$\pi$};

    
    \filldraw[black!90,ultra thick,rounded corners,fill=yellow!30] (9.4,6,0) rectangle (9.9,8,0) node[yellow!90,pos=0.5,rotate=90] (YN512) {FC-512};
    \draw[black,ultra thick,-latex] (9.9,7,0) -- (11,7,0);
    \filldraw[black!90,ultra thick,rounded corners,fill=blue!30] (11,6,0) rectangle (11.9,8,0)  node[black!100,pos=0.5,rotate=90,scale=1.2] {$\psi(s_{t+1})$};
    \node[black,ultra thick,anchor=west] at (12.2,4,0) {Policy Loss};
    \node[black,ultra thick,anchor=west] at (12.2,1,0) (VLN) {Value Loss};
    \node[black,ultra thick,anchor=west] at (12.2,7,0) (SRN) {SR Loss};
    
    \node[gray!70] at (6.7,8,0) (lt1) {$Scene$};
    \node[gray!70,anchor=east] at (14,0,0) (rb1) {e.g. Bathroom 02};
    \begin{scope}[on background layer]
        \node[rounded corners,fill=gray!10, fit=(lt1) (rb1)]{};
        \node[black,draw,ultra thick,rounded corners,fill=purple!20,fit=(A1) (A4) (pi)]{};
    \end{scope} 
\end{tikzpicture}
\end{adjustbox}
    \caption{Reinforcement Learning Network}
    \label{fig:rl}
\end{subfigure}
\caption{Proposed Network Architecture for Hybrid Asynchronous Universal Successor Representations (HAUSR). Network shown in Figure~\ref{fig:rps} is trained first with Algortihm~\ref{alg:1} and the reinforcement learning network in Figure~\ref{fig:rl} is trained with Algorithm~\ref{alg:2}.}
\label{fig:network}
\end{figure*}

%% file: figure1.tex
\begin{wrapfigure}{r}{6.8cm}
\centering
\begin{adjustbox}{width=0.49\textwidth}
\begin{tikzpicture}
    \foreach \X [count=\i] in {69,53,43,37,26}
    {
        \begin{scope}
            \clip [rounded corners=.5cm] ({-(\i*2)+1},0) rectangle coordinate (test1\i) ({-(\i*2)-1},2);
            \node [inner sep=0pt] at (test1\i) {\includegraphics[width=0.13\textwidth]{states/bnw/out_\X}}; 
        \end{scope}

    }
    \node at (-6,2.3) {\texttt{Trained Targets}};
    \node at (-6,-0.7) {\texttt{Successfully Generalized Unseen Targets}};
    \draw[thin,dotted] (-11,-0.3) -- (-1,-0.3);

    \foreach \X [count=\i] in {58,57}
    {
        \begin{scope}
            \clip [rounded corners=.5cm] (-5,{-(\i*2)+1}) rectangle coordinate (test1\i) (-7,{-(\i*2)-1});
            \node [inner sep=0pt] at (test1\i) {\includegraphics[width=0.13\textwidth]{figures/states/bnw/out_\X}}; 
        \end{scope}
    }
    \foreach \X [count=\i] in {47,46}
    {
        \begin{scope}
            \clip [rounded corners=.5cm] (-3,{-(\i*2)+1}) rectangle coordinate (test1\i) (-1,{-(\i*2)-1});
            \node [inner sep=0pt] at (test1\i) {\includegraphics[width=0.13\textwidth]{figures/states/bnw/out_\X}};
        \end{scope}
    }  
    \foreach \X [count=\i] in {44,42}
    {
        \begin{scope}
            \clip [rounded corners=.5cm] (-5,{-(\i*2)+1}) rectangle coordinate (test1\i) (-3,{-(\i*2)-1});
            \node [inner sep=0pt] at (test1\i) {\includegraphics[width=0.13\textwidth]{figures/states/bnw/out_\X}}; 
        \end{scope}
    }  
    \foreach \X [count=\i] in {40,27}
    {
        \begin{scope}
            \clip [rounded corners=.5cm] (-7,{-(\i*2)+1}) rectangle coordinate (test1\i) (-9,{-(\i*2)-1});
            \node [inner sep=0pt] at (test1\i) {\includegraphics[width=0.13\textwidth]{figures/states/bnw/out_\X}}; 
        \end{scope}
    } 
    \foreach \X [count=\i] in {23,22}
    {
        \begin{scope}
            \clip [rounded corners=.5cm] (-9,{-(\i*2)+1}) rectangle coordinate (test1\i) (-11,{-(\i*2)-1});
            \node [inner sep=0pt] at (test1\i) {\includegraphics[width=0.13\textwidth]{figures/states/bnw/out_\X}}; 
        \end{scope}
    } 
\end{tikzpicture}
\end{adjustbox}
\caption{Starting from random states, our agent was trained to navigate to only five trained images. When directed to navigate to the over 19 other targets, our agent was successful in completing these new tasks without any additional training (only some successful targets are shown above). Even though some states had only minor differences, the agent was successful in differentiating and reaching both goals.}
~\label{fig:overview}
\end{wrapfigure}

%% file: algorithms.tex
 \begin{algorithm}
\caption{State Representation and Reward Prediction Learning}
\label{alg:1}
\begin{algorithmic}[1]
\State Initialize transition history $\mathcal{H}$
\For{$g$ training goals}
\State Collect random rollouts  $\mathcal{H}\leftarrow\mathcal{H}\cup (g,s_1,a_1,r_1,s_2,...,s_N,a_N,r_N)$
\EndFor
\For{time step $t$ in $\mathcal{H}$}
\State Pick random transition $(g,s_t,a_t,r_t,s_{t+1},a_{t+1},r_{t+1})$ from $\mathcal{H}$ 
\State Calculate autoencoder, forward and inverse dynamics losses as $\mathcal{L}_{\phi}$ 
\State Perform gradient decent on  $\mathcal{L}_{\phi}$ to update state representation network parameters
\State Perform gradient decent on $ \mathcal{L}_\omega = [r_t - \phi(s_{t+1})^{\top}\omega(g;\theta_\omega)]^2 $ w.r.t. $\theta_\omega$
\EndFor
\end{algorithmic}
\end{algorithm}

\begin{algorithm}
\caption{Async-USR}
\label{alg:2}
\begin{algorithmic}[1]
\For{$\mathcal{M}$ agents simultaneuously}
\EndFor
\For{$ns$ steps} 
\State Obtain rollout $(g,s_1,a_1,r_1,s_2,...,s_N,a_N,r_N)$ by following $\pi(s_t)$
\State Compute $L_\psi = \|\phi(s_t)+\gamma_t\psi(s_{t+1},g;\theta_\psi) - \psi(s_t,g;\theta_\psi)\|^2$
\State Compute $L_V = \|r(s_t)+\gamma_t V(s_{t+1},g;\theta_V) - V(s_t,g;\theta_V)\|^2$
\State Compute $\hat{A}_t^{\psi} = [\phi(s_t)+\gamma_t\psi(s_{t+1},g)-\psi(s_t,g)]^\top \omega_g$
\State Compute $\hat{A}_t^V = \sum_{k=0}^{\infty} \gamma^k r_{t+k} - V_t $
\State Perform gradient descent on loss $(\lambda  \hat{A}_t^{\psi}+ \hat{A}_t^V )log(\pi(s_t,g;\theta_\pi) + 0.5L_V+ 0.1 L_\psi - \beta H(\pi)$ w.r.t. $\theta_\pi,\theta_V$ where $H(\pi)$ is the entropy loss and $\lambda \& \beta$ are hyperparamters
\EndFor

\end{algorithmic}
\end{algorithm}